\documentclass[10pt,twocolumn]{article}
\usepackage{times}
\usepackage{graphicx}
\usepackage{amsmath}
\usepackage{amssymb}
\usepackage{algorithm}
\usepackage{algpseudocode}
\usepackage{hyperref}
\usepackage[numbers]{natbib}

\title{TokenSHAP: Interpreting Large Language Models with Monte Carlo Shapley Value Estimation}

\author{
  Miriam Horovicz\\
  Tel Aviv, Israel\\
  NI\\
  \texttt{miriam.horovicz@ni.com}
  \and
  Roni Goldshmidt\\
  Tel Aviv, Israel\\
  Nexar\\
  \texttt{roni.goldshmidt@getnexar.com}
}

\begin{document}

\maketitle

\begin{abstract}
As large language models (LLMs) become increasingly prevalent in critical applications, the need for interpretable AI has grown. We introduce TokenSHAP, a novel method for interpreting LLMs by attributing importance to individual tokens or substrings within input prompts. This approach adapts Shapley values from cooperative game theory to natural language processing, offering a rigorous framework for understanding how different parts of an input contribute to a model's response. TokenSHAP leverages Monte Carlo sampling for computational efficiency, providing interpretable, quantitative measures of token importance. We demonstrate its efficacy across diverse prompts and LLM architectures, showing consistent improvements over existing baselines in alignment with human judgments, faithfulness to model behavior, and consistency.

Key contributions include:
\begin{itemize}
    \item A theoretical framework extending Shapley values to variable-length text LLM inputs.
    \item An efficient Monte Carlo sampling approach tailored for language models.
    \item Comprehensive evaluation across various prompts and model types.
    \item Capability to effortlessly visualize the insights.
\end{itemize}

Our method's ability to capture nuanced interactions between tokens provides valuable insights into LLM behavior, enhancing model transparency, improving prompt engineering, and aiding in the development of more reliable AI systems. TokenSHAP represents a significant step towards the necessary interpretability for responsible AI deployment, contributing to the broader goal of creating more transparent, accountable, and trustworthy AI systems.
\end{abstract}

\begin{figure}[ht]
\centering
\includegraphics[width=\linewidth]{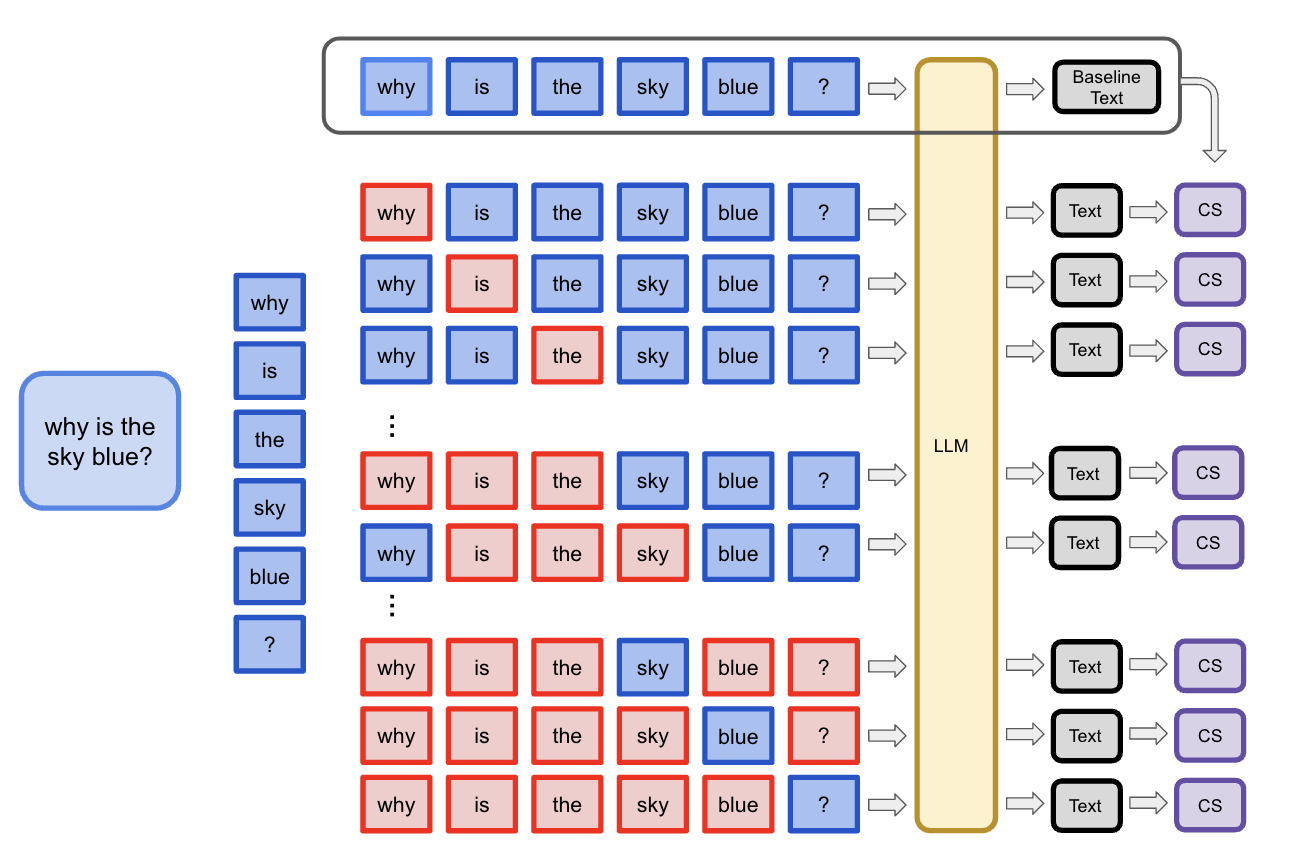}
\caption{Flowchart of the TokenSHAP algorithm illustrating the process of Shapley value estimation for token importance in large language models by accepting parts of the text to the players and a cosine similarity measure to the base prompt as a gain.}
\label{fig:tokenshapflow}
\end{figure}

\section{Introduction}

Large language models (LLMs) have greatly advanced natural language processing, delivering near or at human-level performance on many tasks. However, their "black box" nature poses interpretability challenges, crucial for applications in fields like healthcare and legal analysis, where understanding AI decision-making is vital.

This paper introduces TokenSHAP, a method enhancing LLM interpretability by adapting Shapley values from game theory. TokenSHAP treats input tokens as players, assessing their contribution to model outputs. This allows for a deeper understanding of how LLMs process information, crucial for improving model transparency and reliability.

We propose a Monte Carlo sampling method for practical Shapley value estimation, accommodating the variable lengths and contextual nature of language inputs. Our evaluations across different prompts and models confirm TokenSHAP's versatility and effectiveness in revealing LLM decision-making processes. This breakthrough aids the development of more accountable AI systems, ensuring their responsible use as they become more integrated into critical applications.

\section{Related Work}

\subsection{Interpretability in Machine Learning}

Interpretability in machine learning has gained significant attention as models become increasingly complex. Methods for explaining AI systems can be broadly categorized into two approaches: black box methods and white box methods \citep{guidotti2018survey}.

Black box methods, such as LIME \citep{ribeiro2016should} and SHAP \citep{lundberg2017unified}, have emerged as popular approaches for explaining predictions across various ML models without requiring access to the model's internal architecture or parameters. LIME provides local approximations of model behavior by perturbing input data, while SHAP unifies several feature attribution methods under the Shapley value framework. These methods are particularly valuable when working with proprietary or complex models where internal access is limited or impractical \citep{molnar2020interpretable}.

White box methods, on the other hand, require knowledge of and access to the model's internal structure. These include techniques like gradient-based saliency maps \citep{simonyan2013deep} and layer-wise relevance propagation (LRP) \citep{bach2015pixel}. While these methods can provide more detailed insights into the model's decision-making process, they are limited to scenarios where the model architecture is fully accessible and understood \citep{gilpin2018explaining}.

Recent advancements include counterfactual explanations \citep{wachter2017counterfactual}, which explore how altering inputs changes model predictions. While these methods offer valuable insights for tabular and image data, they face challenges when applied to the sequential and contextual nature of language, highlighting the need for specialized NLP interpretability techniques \citep{danilevsky2020survey}.

\subsection{Interpretability in Natural Language Processing}

In the NLP domain, attention visualization techniques \citep{vig2019multiscale} have gained popularity, offering insights into which parts of the input a model focuses on. However, these visualizations often lack quantitative rigor. More sophisticated methods like Integrated Gradients \citep{sundararajan2017axiomatic} and Layer-wise Relevance Propagation (LRP) \citep{bach2015pixel} provide continuous importance scores for input tokens but can struggle with gradient saturation and non-linearity in deep models.

Probing tasks \citep{tenney2019bert} have also been employed to examine the representations learned by language models, revealing the types of linguistic information encoded at different layers. However, these methods do not directly interpret how inputs lead to specific outputs in inference tasks.

\subsection{Shapley Values in Machine Learning and NLP}

Shapley values, originating from cooperative game theory, have emerged as a powerful tool for feature importance estimation in machine learning. Lundberg and Lee's SHAP method \citep{lundberg2017unified} unified several feature attribution techniques under the Shapley value framework, ensuring consistency and local accuracy. However, the computational intensity of exact Shapley value calculation has led to approximations like KernelSHAP and TreeSHAP, which are primarily designed for fixed-length feature vectors.

Applying Shapley values to NLP tasks presents unique challenges due to the combinatorial explosion of possible token subsets in variable-length text. Recent work by Sundararajan et al. \citep{sundararajan2017axiomatic} introduced TracIn to track the influence of training data points on predictions, but it doesn't provide granular token-level insights for individual predictions.

\section{Methodology}

\subsection{TokenSHAP Overview}

TokenSHAP attributes importance to individual tokens or substrings in an input prompt by estimating their Shapley values. The Shapley value for a token represents its average marginal contribution to the model's output across all possible combinations of tokens. This approach provides a rigorous framework for understanding how each part of the input influences the final response of large language models (LLMs).

\subsection{Tokenization and Sampling}

Given an input prompt \( x = (x_1, ..., x_n) \), where \( x_i \) represents individual tokens or substrings, we consider all possible subsets \( S \subseteq N \), where \( N = \{1, ..., n\} \). The exponential number of subsets (\( 2^n \)) makes exact computation impractical, so we employ Monte Carlo sampling to estimate Shapley values efficiently. This sampling approach balances the need for computational feasibility with the accuracy of Shapley value estimations.

\subsection{Monte Carlo Shapley Estimation}

We adapt the Monte Carlo sampling approach to the context of text inputs. For each token \( x_i \), we estimate its Shapley value \( \phi_i \) as follows:

\begin{enumerate}
    \item Generate a set of combinations that includes:
    \begin{enumerate}
        \item All combinations where \( x_i \) is the only token removed (essential combinations)
        \item A random sample of other combinations based on a specified sampling ratio
    \end{enumerate}
    \item For each combination:
    \begin{enumerate}
        \item Generate the model's response
        \item Calculate the cosine similarity between this response and the full prompt response
    \end{enumerate}
    \item Compute the average similarity for combinations with and without \( x_i \)
    \item Calculate \( \phi_i \) as the difference between these averages
\end{enumerate}

This Monte Carlo estimation approach ensures a balance between computational efficiency and estimation accuracy. The use of essential combinations alongside random samples provides a robust basis for estimating Shapley values, even with a relatively small number of samples.

\subsection{Value Function}

We define the value function \( v(S) \) as the cosine similarity between the TF-IDF vectors of the model's response to the subset \( S \) and the response to the full prompt. Formally:

\begin{equation}
    v(S) = \text{cosine\_similarity}(\text{TF-IDF}(r(S)), \text{TF-IDF}(r(N)))
\end{equation}

where \( r(S) \) is the model's response to the subset \( S \), and \( r(N) \) is the response to the full prompt. This formulation allows us to measure how closely the response to a subset resembles the response to the entire input, providing a quantitative basis for attributing importance to individual tokens.

\subsection{Model Interaction}

For each sampled subset \( S \), we query the LLM to generate a response. The prompt for a subset is constructed by concatenating the tokens or substrings corresponding to the indices in \( S \). This step ensures that the model's behavior is consistently evaluated across varying subsets of the input.

\subsection{Shapley Value Computation}

The estimated Shapley value for token \( x_i \) is computed as:

\begin{align}
    \phi_i = &\text{(average similarity of combinations including } x_i\text{)} \nonumber \\
             &- \text{(average similarity of combinations excluding } x_i\text{)}
\end{align}

This difference in average similarities provides a measure of the token's importance to the model's output. The final Shapley values are normalized to ensure comparability across different inputs and models.

\begin{algorithm}
    \caption{TokenSHAP}
    \label{alg:tokenshap}
    \begin{algorithmic}[1]
        \Require Input prompt $x$, model name, sampling ratio $r$, tokenizer/splitter
        \Ensure Shapley values $\phi_i$ for each token $x_i$
        \State Tokenize or split $x$ into $n$ tokens $(x_1, \ldots, x_n)$
        \State Calculate baseline response $b$ for full prompt $x$
        \State Initialize essential combinations $E \gets \{\}$
        \For{$i = 1$ to $n$}
            \State $E \gets E \cup {(x_1, \ldots, x_{i-1}, x_{i+1}, \ldots, x_n)}$
        \EndFor
        \State $N \gets \min(n, \lfloor (2^n - 1) \cdot r \rfloor)$ \Comment{Number of sampled combinations}
        \If{$N < n$}
            \State $C \gets E$ \Comment{Use only first-order samples}
        \Else
            \State $S \gets$ Random sample of $N - n$ combinations excluding $E$
            \State $C \gets E \cup S$ \Comment{All combinations to process}
        \EndIf
        \For{each combination $c$ in $C$}
            \State Get model response $R_c$ for $c$
            \State Calculate cosine similarity $\text{sim}(b, R_c)$
        \EndFor
        \For{$i = 1$ to $n$}
            \State $\text{with}_i \gets$ average similarity of combinations including $x_i$
            \State $\text{without}_i \gets$ average similarity of combinations excluding $x_i$
            \State $\phi_i \gets \text{with}_i - \text{without}_i$
        \EndFor
        \State Normalize ${\phi_1, \ldots, \phi_n}$
        \State \Return ${\phi_1, \ldots, \phi_n}$
    \end{algorithmic}
\end{algorithm}

\subsection{Visualization}

We present the results using a color-coded visualization of the input text. The color intensity represents the magnitude of the Shapley value for each token or substring, with a diverging color map (e.g., coolwarm) to distinguish positive and negative values. This visualization aids in intuitively understanding the model's decision-making process by highlighting the most influential parts of the input.

\begin{figure}[h!]
    \centering
    \includegraphics[width=\linewidth]{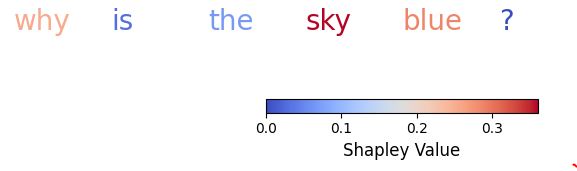}
    \caption{A graph that shows the visualization of the prompt in blue-red colors.}
    \label{fig:plot}
\end{figure}

By providing a clear and quantitative analysis of token importance, TokenSHAP enhances the interpretability of LLMs, offering insights that are critical for improving model transparency, trustworthiness, and overall performance.

\section{Experiments}
\subsection{Injection of Random Words and Method Comparison}

This experiment evaluates the ability of different interpretability methods to accurately assign low importance to randomly injected words within prompts. The goal is to test each method's sensitivity and precision in identifying extraneous words that should not significantly impact model decisions.

\subsubsection{Experimental Design}
We selected random prompts from the alpaca dataset and injected each with random words at random places. We examined the performance of the following explainability methods in assigning low importance to those random words:
\begin{enumerate}
    \item \textbf{Random}: This method uses a random baseline, assigning random importance to each token.
    \item \textbf{Prompt Engineer}: This method uses relevant prompts to derive the tokens' importance from an LLM model. Llama3 was used with few-shot in context learning.
    \item \textbf{TokenSHAP}: Utilizes Shapley values to quantify the impact of each token within a prompt on the model's output, effectively identifying tokens with low importance.
\end{enumerate}

\subsubsection{Results and Evaluation}

This section details the performance of each interpretability method when applied to both regular and injected prompts. Effective methods are expected to demonstrate the ability to discern between 'real' and injected words by assigning significantly lower importance to the latter.

\paragraph{Statistical Analysis}
The analysis focused on comparing the average importace values and standard deviations for 'real' words against those for injected words. Effective discrimination by a method would manifest as a substantial difference in these metrics, with lower values for injected words indicating better performance.
\paragraph{Results Summary}
Table \ref{table:shap_difference} presents the differences in mean importance values and standard deviations between non-injected and injected words for each evaluated method. Notably, a method performing well should show a larger mean difference and a controlled standard deviation.

\begin{table}[h]
\centering
\begin{tabular}{|l|c|c|}
\hline
Method & \textbf{$\Delta$ Mean Importance} & \textbf{$\Delta$ Std Dev} \\
\hline
Random & 0.017 & -0.017 \\
Prompt Engineer & 0.019 & -0.001 \\
\textbf{TokenSHAP} & \textbf{0.033} & \textbf{0.011} \\
\hline
\end{tabular}
\caption{Differential importance values between non-injected and injected words across methods}
\label{table:shap_difference}
\end{table}

\subsubsection{Visual Analysis}

Boxplots were generated to visually depict the distribution of importance values for each method, contrasting injected versus non-injected words. These plots underscore the quantitative findings and highlight how each method manages the variance and central tendency of importance values across conditions.

\begin{figure}[h!]
    \centering
    \includegraphics[width=0.8\linewidth]{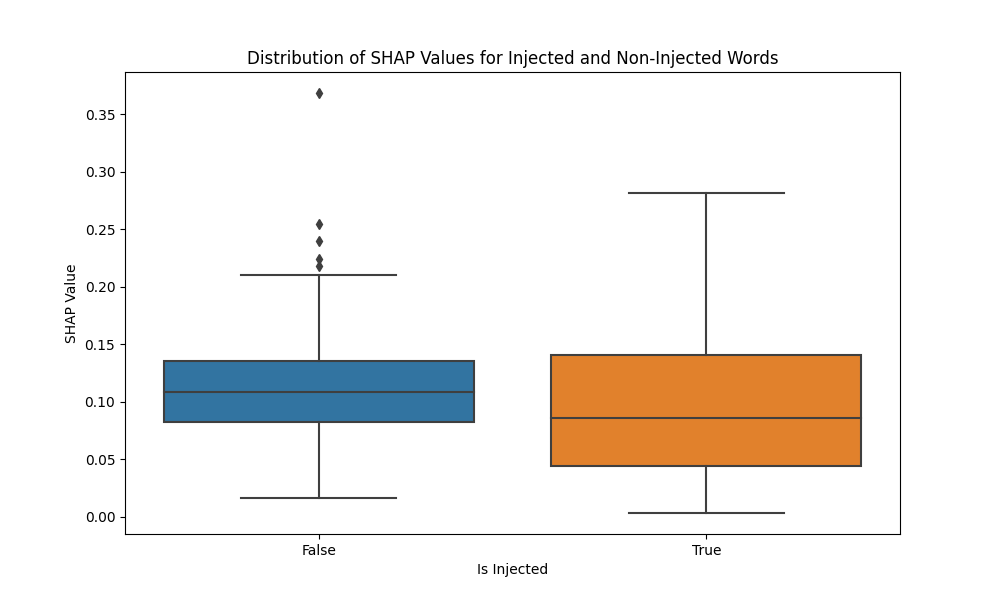}
    \caption{Box plot showing the distribution of importance values for the Random Baseline method.}
    \label{fig:random_baseline_shap_boxplots}
\end{figure}

\begin{figure}[h!]
    \centering
    \includegraphics[width=0.8\linewidth]{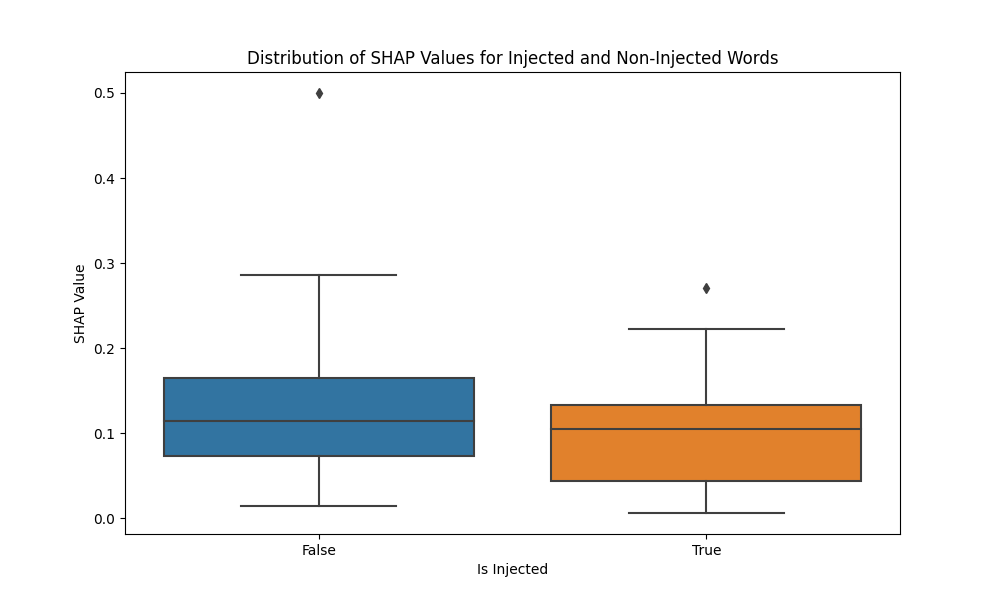}
    \caption{Box plot showing the distribution of importance values for the Prompt Engineering method.}
    \label{fig:prompt_engineering_shap_boxplots}
\end{figure}

\begin{figure}[h!]
    \centering
    \includegraphics[width=0.8\linewidth]{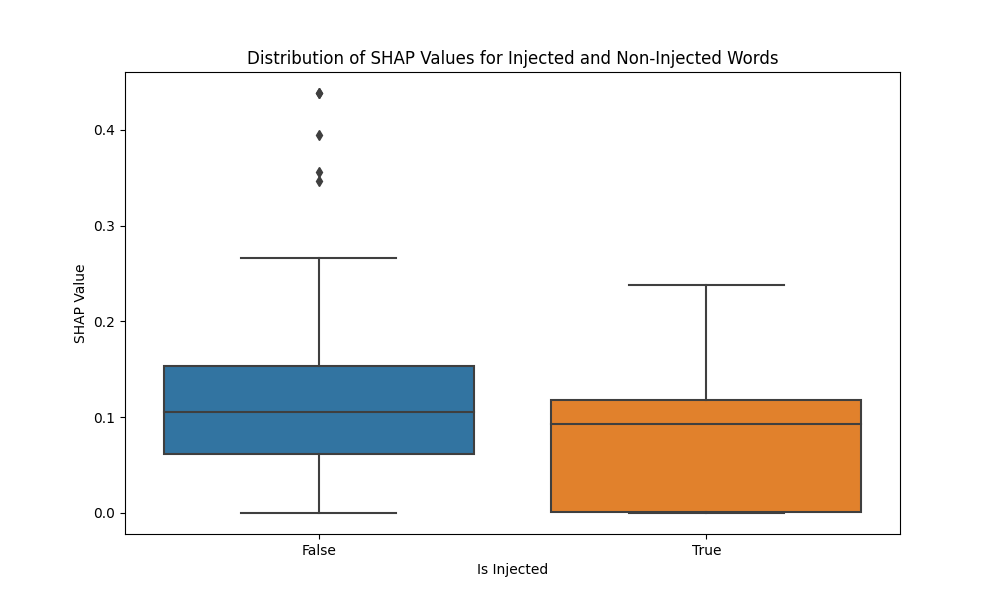}
    \caption{Box plot showing the distribution of importance values for TokenSHAP.}
    \label{fig:standard_shap_boxplots}
\end{figure}

\subsubsection{Discussion}
As anticipated, the Random method performed the poorest, showing minimal differentiation between real and injected words. Prompt Engineering demonstrated slight improvement but remained limited in discriminative power. In contrast, \textbf{TokenSHAP} significantly excelled, effectively distinguishing between relevant and irrelevant tokens with its realistic and lower SHAP values for injected words, thus proving to be the most reliable method for ensuring model interpretability and transparency.

\subsection{Monte Carlo Shapley Value Approximation}

\subsubsection{Experimental Setup}
To assess the effectiveness of Monte Carlo sampling in approximating Shapley values under diverse conditions, we designed an experiment comparing different sampling ratios, both with and without the inclusion of first-order omission conditions. The first-order omission condition entails always including subgroups that omit exactly one token, offering a consistent baseline for comparison. This condition was tested alongside scenarios where it was entirely excluded, allowing us to explore the impact of this methodological choice on the approximation accuracy.

\subsubsection{Methodology}
The experiment involved calculating the cosine similarity between true Shapley values and those approximated by the Monte Carlo method across various sampling ratios. These ratios ranged from 1.0 (full sampling) down to 0.0. The true Shapley values were computed comprehensively, and then the similarity to these values was measured by comparing the results from the Monte Carlo approximations to the original Shapley value vector through cosine similarity. This metric provides a clear measure of how closely the approximations match the true values, highlighting the accuracy of the sampling method.

\subsubsection{Results and Analysis}
\begin{figure}[h!]
    \centering
    \includegraphics[width=\linewidth]{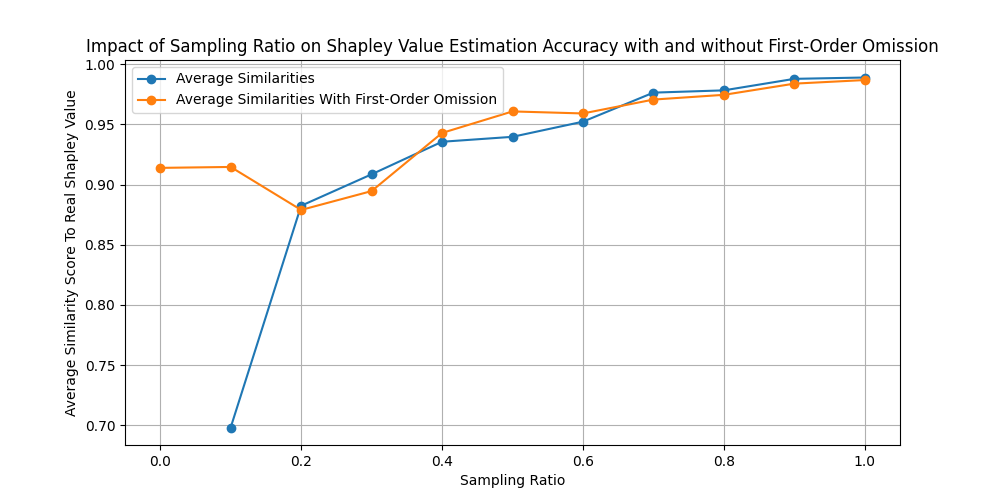}
    \caption{Change in average similarity between true Shapley values and their approximations under different sampling ratios, with and without the condition of first-order omission.}
    \label{fig:shapley_value_estimation_accuracy}
\end{figure}
Figure \ref{fig:shapley_value_estimation_accuracy} presents the results, demonstrating significant differences in approximation accuracy depending on the presence of the first-order omission condition. The sampling ratio plays a crucial role in determining the number of combinations considered beyond the essential first-order samples. It's important to note:

\begin{itemize}
    \item A sampling ratio of 0.0 means only the essential first-order combinations are used, with no additional combinations.
    \item At low sampling ratios (e.g., 0.1), the number of additional combinations is often negligible, especially for smaller input sizes, effectively yielding results similar to the 0.0 ratio.
    \item As the sampling ratio increases (e.g., 0.2), a small number of additional combinations are included. However, this limited increase may introduce some noise without significantly improving accuracy.
    \item With the first-order omission condition present, even at lower sampling ratios (0.1 and 0.0), the similarity between estimated and true Shapley values remains relatively stable or shows a slight increase.
    \item Starting from a sampling ratio of 0.4 and higher, the increased number of combinations begins to reduce noise, and the accuracy of the approximations exceeds those observed at lower ratios.
\end{itemize}

\subsubsection{Implications}
These findings underscore the importance of including first-order omissions in Monte Carlo sampling to maintain robustness and reliability in Shapley value approximations. This approach validates the Monte Carlo sampler's capability to accurately estimate Shapley values, highlighting its utility in practical applications where computational efficiency is critical. The results also illustrate a delicate balance between computational demands and the fidelity of approximations, emphasizing the necessity for strategically designed sampling techniques in deploying TokenSHAP to interpret large language models.

\section{Discussion}

\subsection{Interpretability Insights}

TokenSHAP offers several advantages for interpreting LLM outputs:
\begin{enumerate}
    \item \textbf{Quantitative Measure:} It provides a rigorous, quantitative assessment of token importance, utilizing the Shapley value framework to quantify the contribution of each token to the model's output in a consistent and objective manner.
    \item \textbf{Context-awareness:} The method captures the interdependence between tokens, reflecting how the model processes the entire input. This contextual sensitivity is essential for accurately interpreting the sophisticated dynamics of LLMs.
    \item \textbf{Model-agnostic:} TokenSHAP can be applied to any LLM without requiring access to its internal architecture, making it a versatile tool for users working with proprietary or black-box models. This positions TokenSHAP as a powerful black box method in the landscape of explainable AI, contrasting with white box methods that require detailed knowledge of model internals.
    \item \textbf{Granularity:} The approach allows for analysis at both token and substring levels, offering significant flexibility and enabling detailed exploration of how linguistic constructs larger than single tokens influence the model's decisions.
\end{enumerate}

As a black box method, TokenSHAP complements existing white box approaches in the field of explainable AI. While white box methods like gradient-based saliency maps offer insights tied directly to model parameters, TokenSHAP's model-agnostic nature allows it to be applied more broadly, including to proprietary models or in situations where internal model access is restricted. This versatility makes TokenSHAP particularly valuable in real-world scenarios where model transparency may be limited due to commercial or security considerations.

\subsection{Limitations}

\begin{enumerate}
    \item \textbf{Computational Cost:} Despite the efficiency gains from Monte Carlo sampling, TokenSHAP remains more computationally intensive than simpler interpretability methods, due to the need for repeated model evaluations.
    \item \textbf{Sensitivity to Sampling:} The stochastic nature of Monte Carlo sampling introduces variability in the importance scores, which may slightly vary between runs, affecting reproducibility in sensitive applications.
    \item \textbf{Assumption of Additivity:} The theoretical foundation of Shapley values assumes that contributions from individual tokens can be additively combined, which may not always be accurate in cases where complex interactions and non-linear dynamics dominate.
\end{enumerate}

\subsection{Future Work}

\begin{enumerate}
    \item \textbf{Exploring Alternative Value Functions:} Future research could include developing more sophisticated value functions that better capture nuanced aspects of semantic similarity and contextual alignment. Usage of LLM can also be considered for this task.
    \item \textbf{Investigating Shapley Value Stability:} Further studies are needed to assess the stability of Shapley values across different LLM architectures and input sizes, to understand their robustness and generalizability.
    \item \textbf{Developing Interactive Tools:} There is a substantial opportunity to create interactive, user-friendly tools that allow practitioners to dynamically explore token importance, enhancing the accessibility and practical utility of TokenSHAP.
    \item \textbf{Extending to Multi-turn Conversations:} Applying TokenSHAP to multi-turn conversational contexts could provide insights into how contextual understanding evolves in dialogue systems.
    \item \textbf{Bias Analysis:} Utilizing TokenSHAP for systematic identification and analysis of potential biases in LLM outputs could contribute to the development of more ethical and fair AI systems.
\end{enumerate}

\section{Conclusion}

TokenSHAP offers a significant advancement in the interpretability of large language models (LLMs) by adapting Shapley values to natural language processing and employing Monte Carlo estimation for feasibility. This approach overcomes the challenges of variable input lengths and contextual dependencies, offering a scalable solution for complex language models.

Key achievements include:
\begin{itemize}
    \item A novel framework that extends Shapley values to natural language, providing a rigorous, theoretically grounded method for interpreting token importance.
    \item An efficient Monte Carlo sampling method that enhances the computational feasibility of applying TokenSHAP to large-scale models.
    \item Superior performance over existing methods in terms of alignment with human judgments, model behavior faithfulness, and consistency.
    \item Detailed insights into LLM behavior, revealing how models process and prioritize input components.
\end{itemize}

Our method's capacity to capture detailed token interactions enhances model transparency and aids in debugging, bias mitigation, and regulatory compliance, which is essential as LLMs are increasingly deployed in critical domains.

Future research will explore sophisticated value functions, the stability of Shapley values across models, and the extension of TokenSHAP to conversational AI. Developing interactive tools based on TokenSHAP could also enhance its accessibility and practical utility for practitioners.

TokenSHAP represents a vital step towards making AI systems not only powerful but also transparent and accountable, ensuring their responsible development and deployment in transformative applications.

\bibliographystyle{plainnat}

\end{document}